\documentclass[11pt]{article}

\usepackage{ACL2023}

\usepackage{times}
\usepackage{latexsym}

\usepackage[T1]{fontenc}
\usepackage[utf8]{inputenc}

\usepackage{microtype}

\usepackage{inconsolata}
\usepackage{graphicx}

\title{BLAD: A Historically Contextualized, Multilingual Dataset of Bangladeshi Legal Acts (1799–2025)}

\author{Adib Sakhawat \\
  Systems and Software Lab (SSL) \\
  Department of Computer Science and Engineering\\
  Islamic University of Technology, Dhaka, Bangladesh \\
  \texttt{sakhadib@gmail.com}}

\begin{document}
\maketitle
\begin{abstract}
We present the Bangladesh Legal Acts Dataset (BLAD), a curated collection of 1{,}484 legislative acts enacted between 1799 and 2025. Each act is represented with its full text, structured sections and footnotes, repeal status, and metadata linking it to the governing regime, head of state, and prevailing legal framework at the time of enactment. The corpus spans English, Bengali, and mixed-language documents, supporting temporal and multilingual analysis of statutory law. BLAD addresses a persistent gap in legal natural language processing (NLP) resources for low-resource, civil-law jurisdictions in South Asia. We describe the acquisition and enrichment pipeline, report descriptive statistics over more than two centuries of legislation, and outline the research directions the corpus enables. The dataset is publicly available under the CC~BY-SA~4.0 license at \url{https://www.kaggle.com/datasets/sakhadib/bangladesh-legal-acts-dataset}.
\end{abstract}

\section{Introduction}

Legal natural language processing (NLP) has advanced rapidly in recent years, driven by the digitisation of legal corpora and by progress in transformer-based architectures. This growth, however, has been distributed unevenly across jurisdictions: large, well-curated legal datasets overwhelmingly concern English-language common-law systems, while civil-law frameworks and non-English legal traditions remain comparatively underserved \cite{niklaus_lextreme_2023}. The imbalance constrains the development of inclusive legal AI systems and limits cross-jurisdictional legal research.

Bangladesh presents an instructive case of this gap. Its population of roughly 180 million is governed by a hybrid legal system that has accumulated across more than two centuries of colonial, post-partition, and post-independence rule. Despite recent efforts to build legal datasets for neighbouring South Asian jurisdictions \cite{k_dataset_2025}, the Bangladeshi statutory corpus has remained largely inaccessible to computational researchers, owing to inconsistent digitisation, mixed-language documentation, and the absence of structured, machine-readable formats.

Existing Bangladeshi legal NLP resources share three limitations. First, they are narrow in scope: the largest English-only corpus comprises 595 acts and 18{,}023 sections \cite{wasi_exploring_2024}, less than 40\% of the complete statutory framework. Second, they lack historical contextualisation, offering no systematic link between an act and the governmental period in which it was enacted, which precludes temporal legal analysis. Third, their focus on English-only content overlooks the bilingual character of Bangladeshi law, in which Bengali and English coexist in patterns that reflect colonial transitions and contemporary language policy.

\begin{figure}[!htb]
    \centering
    \includegraphics[width=1\linewidth]{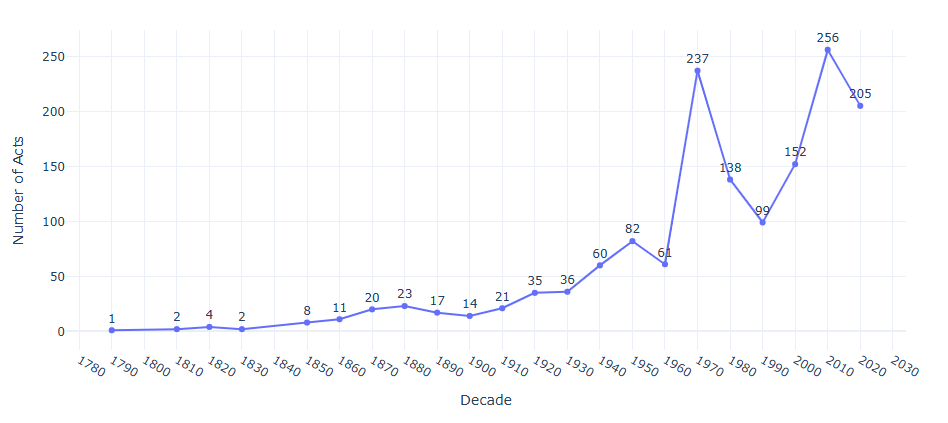}
    \caption{Number of acts per decade across the full temporal span of the corpus.}
    \label{fig:acts-per-decade}
\end{figure}

These limitations carry practical consequences. Bangladesh's judiciary faces a backlog of several million pending cases, which creates a concrete demand for automated legal information systems. The absence of comprehensive, structured statutory corpora impedes the development of retrieval-augmented generation systems, legal summarisation tools, and related applications that could broaden access to legal information.

This paper introduces the \textbf{Bangladesh Legal Acts Dataset} (BLAD), a corpus that addresses these gaps through three contributions:

\textbf{Scale and coverage.} BLAD comprises 1{,}484 legislative acts spanning 226 years (1799--2025), with 35{,}633 sections and 14{,}523 footnotes, together with section-level language identification. This is an approximately 2.5$\times$ increase over the largest previously available dataset and provides substantially complete coverage of the statutory framework.

\textbf{Historical contextualisation.} Each act is enriched with metadata describing the governing regime at the time of enactment, spanning fourteen distinct government systems and forty-two heads of state, from Company Rule (1799--1858) to the contemporary period. This layer enables research on temporal legal evolution and regime-specific legislative behaviour.

\textbf{Multilingual and structural annotation.} Every act is segmented into sections and footnotes and annotated with a detected language label (English, Bengali, or mixed) and a token count, providing structured supervision for multilingual and document-structure tasks.

The dataset is released under the CC~BY-SA~4.0 license to support reproducible research and extension to other South Asian legal systems.\footnote{Dataset: \url{https://www.kaggle.com/datasets/sakhadib/bangladesh-legal-acts-dataset} (DOI: \href{https://doi.org/10.34740/KAGGLE/DSV/12511542}{10.34740/KAGGLE/DSV/12511542}).}

\section{Related Work}

Legal NLP has grown substantially over the past decade, driven by the digitisation of legal systems and by advances in transformer-based architectures \cite{niklaus_lextreme_2023}. Recent surveys document a shift from early rule-based approaches to large language models tailored for legal text \cite{ariai_natural_2024}. The distinctive properties of legal text---long documents, complex syntax, domain-specific terminology, and a stringent requirement for accuracy and reliability---have motivated specialised methods that diverge from general-domain NLP \cite{ariai_natural_2024}.

India has led regional efforts through comprehensive benchmarks. IL-TUR (Indian Legal Text Understanding and Reasoning) \cite{joshi_il-tur_2024} defines eight tasks covering English and nine Indian languages. IndicLegalQA \cite{k_dataset_2025} contributes 10{,}000 question--answer pairs drawn from 1{,}256 Indian Supreme Court judgments, addressing the need for legal information retrieval in a context where a backlog exceeding 43 million cases motivates automated processing \cite{k_dataset_2025}.

By contrast, the South Asian legal NLP landscape has remained largely incomplete with respect to Bangladesh. Earlier Bangladesh-specific work was confined to preliminary efforts such as legal-text visualisation systems \cite{mandal_visual_2017} and narrow domain-specific applications, rather than comprehensive coverage of the legislative framework.

East Asian jurisdictions, particularly China and South Korea, have established leadership in large-scale legal dataset construction and systematic benchmarking \cite{xiao_cail2018_2018}. The Chinese AI and Law Challenge (CAIL) dataset, introduced in 2018, comprised 2.6 million criminal cases from the Supreme People's Court of China. Subsequent Chinese benchmarks such as LexEval \cite{li_lexeval_2024} span 23 tasks and 14{,}150 questions across diverse legal-reasoning abilities. South Korea has contributed LBOX~OPEN \cite{hwang_multi-task_2022}, comprising 147{,}000 legal precedents across tasks including judgment prediction, case classification, and summarisation; the Korean Benchmark for Legal Language understanding (KBL) \cite{kim_developing_2024} extends this line with pragmatic evaluation designed for large language models.

Standardised evaluation frameworks have been central to the field's reproducibility. LexGLUE \cite{chalkidis_lexglue_2022} established English-language evaluation standards across seven legal tasks. Building on it, LEXTREME \cite{niklaus_lextreme_2023} targets multilingual evaluation, incorporating 11 datasets across 24 languages and 17 jurisdictions. LegalBench \cite{guha_legalbench_2023} advances collaborative benchmark construction through 162 tasks developed jointly by legal professionals and NLP researchers.

Recent dataset construction has emphasised systematic collection, annotation, and quality assurance, reflecting a maturation from ad hoc web scraping to structured corpus curation. MultiLegalSBD \cite{brugger_multilegalsbd_2023} illustrates the difficulty of multilingual sentence-boundary detection in legal preprocessing, and specialised resources demonstrate the value of expert annotation for regulatory applications \cite{joshi_il-tur_2024}.

Several gaps persist despite this progress. First, most large-scale legal datasets concern common-law jurisdictions, particularly those using English as the primary legal language \cite{niklaus_lextreme_2023}. Second, existing multilingual datasets often lack the historical depth and systematic contextualisation required for the study of legal evolution, concentrating instead on contemporary documents \cite{niklaus_multilegalpile_2024}. Third, the prevalence of case-law and judicial-decision datasets has left statutes, regulations, and legislative documents comparatively underrepresented, despite their foundational role \cite{niklaus_lextreme_2023}.

These gaps motivate the present work. Integrating historical government context with statutory text addresses the shortfall in temporal analysis capabilities, and constructing a comprehensive civil-law resource addresses the underrepresentation of such systems in current multilingual benchmarks.

\section{Dataset Description}

BLAD is a comprehensive collection of legislative acts from Bangladesh spanning 1799 to 2025. It was constructed to address the fragmented and lightly processed state of Bangladeshi legal data, which has impeded large-scale legal NLP. The corpus provides a structured, machine-readable representation of the known statutory framework, enriched with contextual information about the governments and legal systems under which each act was enacted.

\subsection{Overview}
The corpus contains 1{,}484 legislative acts, 35{,}633 sections, and 14{,}523 footnotes. Acts appear in English, Bengali, and mixed-language forms, making the corpus suitable for multilingual legal NLP. Each act is stored in a standardised JSON record carrying its title, year, language, token count, and links to historical government and legal-system context.

\subsection{Data Sources and Processing}
\begin{figure}[!htb]
    \centering
    \includegraphics[width=1\linewidth]{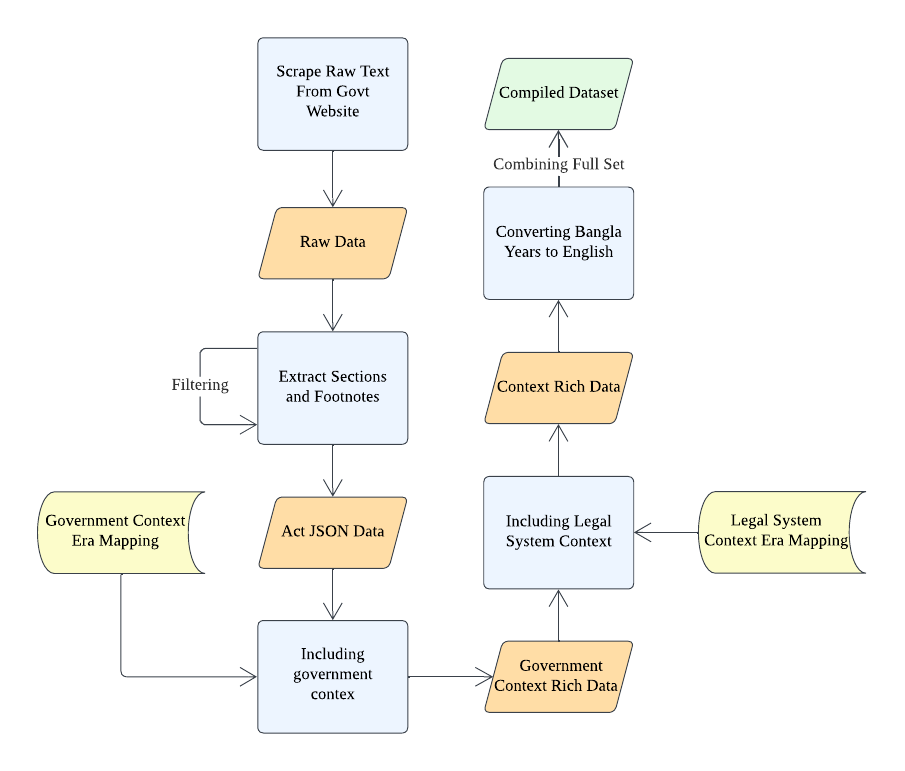}
    \caption{The six-stage acquisition and enrichment pipeline used to construct BLAD.}
    \label{fig:pipeline}
\end{figure}

The corpus was compiled from the official Bangladesh Laws Portal maintained by the Ministry of Law, Justice and Parliamentary Affairs. The construction pipeline (Figure~\ref{fig:pipeline}) proceeds through six stages, each addressing a specific source of noise or a specific enrichment objective.

\paragraph{Stage 1: Acquisition.} An index crawler enumerated the act listing and, for each act, recorded its canonical source URL, act number, title, and year of enactment. Requests were rate-limited with a fixed delay between successive calls to avoid placing undue load on the portal, and failures were logged for retry so that individual errors did not interrupt the crawl.

\paragraph{Stage 2: Structural parsing.} Each act page was parsed into its constituent sections and footnotes. Because the portal's markup is inconsistent across acts of different eras, parsing relied on the recurring structural cues that delimit a section body and its associated footnotes, rather than on a single fixed template. Sections and footnotes were preserved as ordered lists so that document structure is recoverable downstream.

\paragraph{Stage 3: Normalisation and cleaning.} Extracted text was stripped of residual markup and boilerplate, and inconsistent fields were standardised to a common schema. Bengali numerals appearing in years and identifiers were mapped to their Western Arabic equivalents to make temporal fields sortable, and malformed or duplicated fragments were repaired or removed.

\paragraph{Stage 4: Language identification.} Each section was assigned a language label. Because Bengali and Latin scripts occupy disjoint Unicode ranges, a script-ratio heuristic distinguishes predominantly Bengali from predominantly English text, and labels a section as \emph{mixed} when both scripts occur above a threshold. Token counts were computed for each act using a regular-expression tokeniser applied to the cleaned text.

\paragraph{Stage 5: Historical and legal-system contextualisation.} Each act was linked to the political period in which it was enacted by matching its year against a curated lookup table of governmental periods---for example Company Rule, British Colonial Rule, the Pakistan period, and post-independence democratic and military regimes---and to the corresponding government and head of state. A parallel mapping records the broader legal-system context of the period, including the governing legal framework, court structure, policing arrangements, and land-relations regime. This contextual layer is derived from secondary historical sources rather than from the acts themselves, and is therefore treated as an interpretive annotation rather than as primary data.

\paragraph{Stage 6: Validation and recovery.} Combined records were validated for structural consistency, and an automated recovery step re-derived context for acts whose metadata was initially missing or corrupted. Processing statistics, source URLs, and timestamps were retained with each record to support auditing and reproducibility.

\subsection{Dataset Features}
Each act carries the following fields:
\begin{itemize}
    \item \textbf{Act title:} the title of the legislative act.
    \item \textbf{Act year:} the year of enactment.
    \item \textbf{Language:} the language(s) used in the act (English, Bengali, or mixed).
    \item \textbf{Token count:} the number of tokens in the act, computed via regular-expression tokenisation.
    \item \textbf{Sections:} an ordered list of sections with their content.
    \item \textbf{Footnotes:} an ordered list of footnotes, including amendment history where present.
    \item \textbf{Repeal status:} whether the act is currently repealed.
    \item \textbf{Government context:} the government in power at enactment, including its name, head of government, period of rule, and mode of accession.
    \item \textbf{Legal-system context:} the legal framework, period, and policing arrangements in force at enactment.
    \item \textbf{Processing information:} the tokenisation method, source URL, and metadata-recovery status.
\end{itemize}

The corpus is intended as a foundational resource for research on Bangladeshi legal history, multilingual legal NLP, and legal-technology applications in the Global South.

\section{Dataset Statistics and Analysis}

BLAD supports descriptive analysis of more than two centuries of legislative activity. This section reports statistics on the volume of acts, legislative trends over time, the distribution of activity across government systems and leaders, and the structural complexity of individual acts. The trends reported here are descriptive; we make no causal claims, and readers should bear in mind that archival coverage may be uneven across periods.

\subsection{Corpus Overview}
The corpus comprises the following:
\begin{itemize}
    \item \textbf{Total acts:} 1{,}484
    \item \textbf{Total sections:} 35{,}633
    \item \textbf{Total footnotes:} 14{,}523
    \item \textbf{Languages:} English, Bengali, mixed
    \item \textbf{Time span:} 1799--2025
    \item \textbf{Government systems:} 14
    \item \textbf{Heads of state:} 42
    \item \textbf{Token count:} over 2.5 million tokens
\end{itemize}
These figures span approximately 226 years of legal history across several distinct political eras.

\subsection{Legislative Activity by Decade}
\begin{figure}[!htb]
    \centering
    \includegraphics[width=1\linewidth]{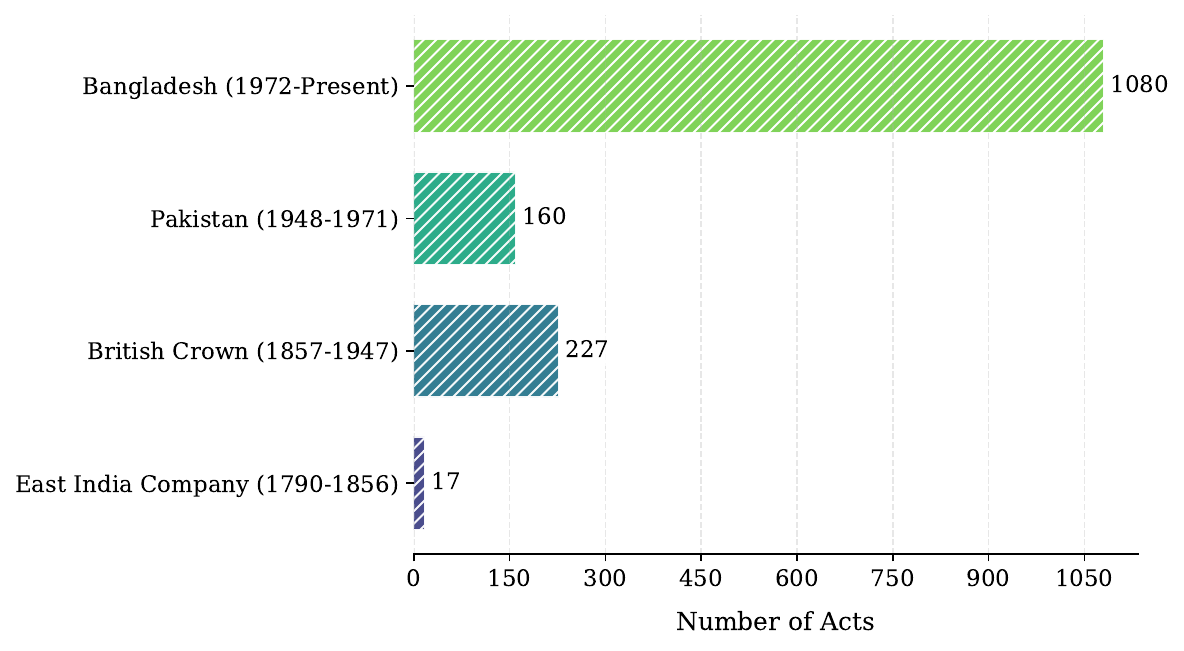}
    \caption{Distribution of legal acts by historical period.}
    \label{fig:period-distribution}
\end{figure}

Legislative output has increased markedly over time, with pronounced changes around major political transitions (Figure~\ref{fig:period-distribution}).

\begin{itemize}
    \item \textbf{1790s--1850s (Company Rule):} fewer than ten acts per decade, establishing the foundational colonial legal framework.
    \item \textbf{1860s--1950s (British Colonial Rule):} an average of 11--82 acts per decade, indicating steady growth.
    \item \textbf{1950s--1970s (Pakistan period):} a transitional phase accounting for 72 acts (4.9\% of the total).
    \item \textbf{1970s (post-independence):} a sharp rise in activity, with 237 acts enacted.
    \item \textbf{2010s:} the most productive decade, with 256 acts.
    \item \textbf{2020s:} sustained high activity, with 205 acts to date.
\end{itemize}

The acceleration of legislation from the 1970s onward is consistent with the growing administrative complexity of the post-independence state.

\subsection{Government Systems and Legislative Productivity}
\begin{figure}[!htb]
    \centering
    \includegraphics[width=1\linewidth]{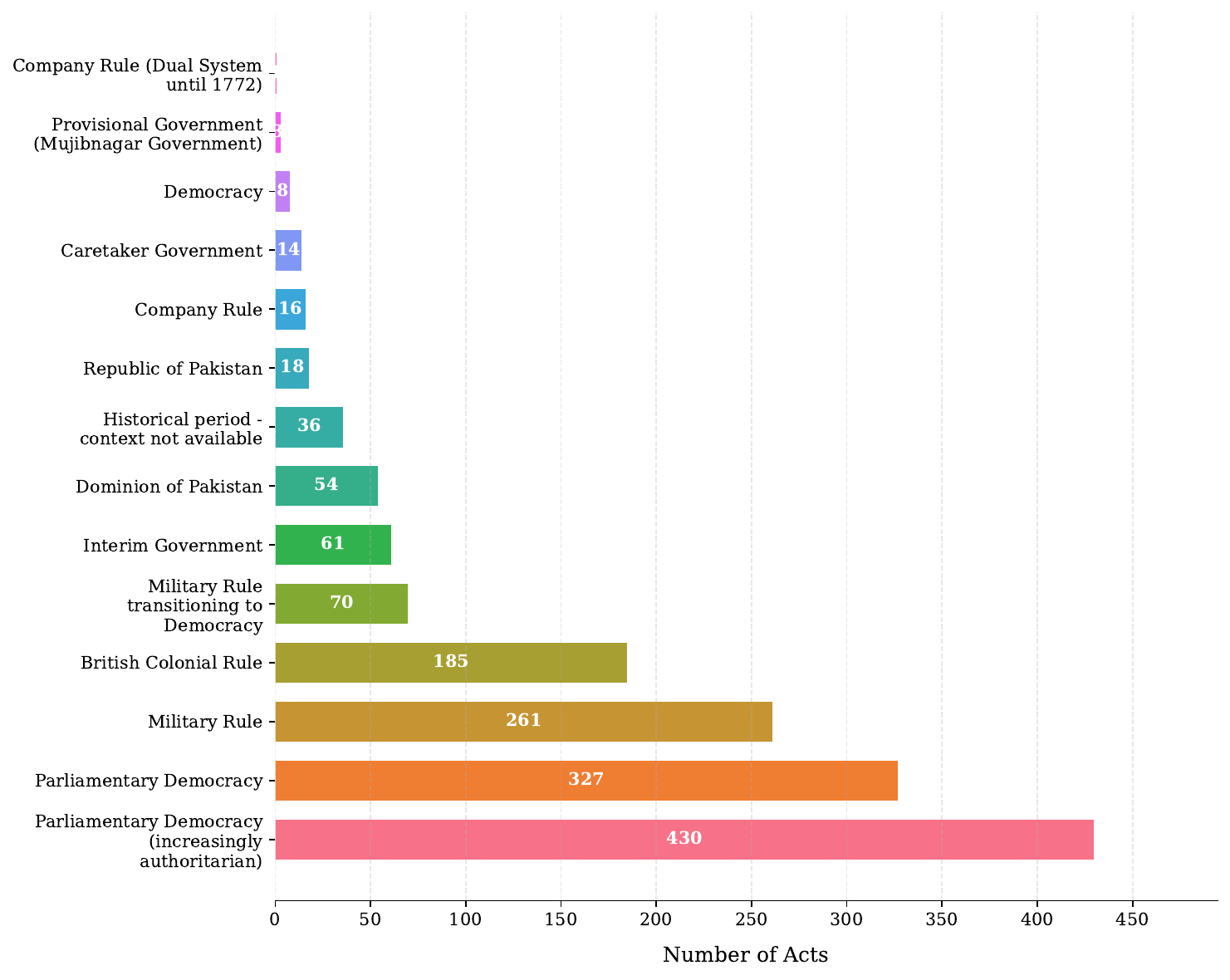}
    \caption{Distribution of legal acts by government system.}
    \label{fig:govt-distribution}
\end{figure}

Legislative output varies substantially across government systems (Figure~\ref{fig:govt-distribution}).

\begin{itemize}
    \item \textbf{Parliamentary democracy (increasingly authoritarian):} 430 acts, followed by regular parliamentary democracy with 327 acts; together these account for over half of all acts.
    \item \textbf{Military rule:} 261 acts, concentrated on administrative reorganisation and policy implementation.
    \item \textbf{British colonial rule:} 185 acts, primarily establishing colonial administrative frameworks.
\end{itemize}

Democratic periods are associated with higher legislative volume than authoritarian ones in this corpus; we report this as an observed association rather than a causal effect.

\subsection{Leadership and Legislative Productivity}
\begin{figure}[!htb]
    \centering
    \includegraphics[width=1\linewidth]{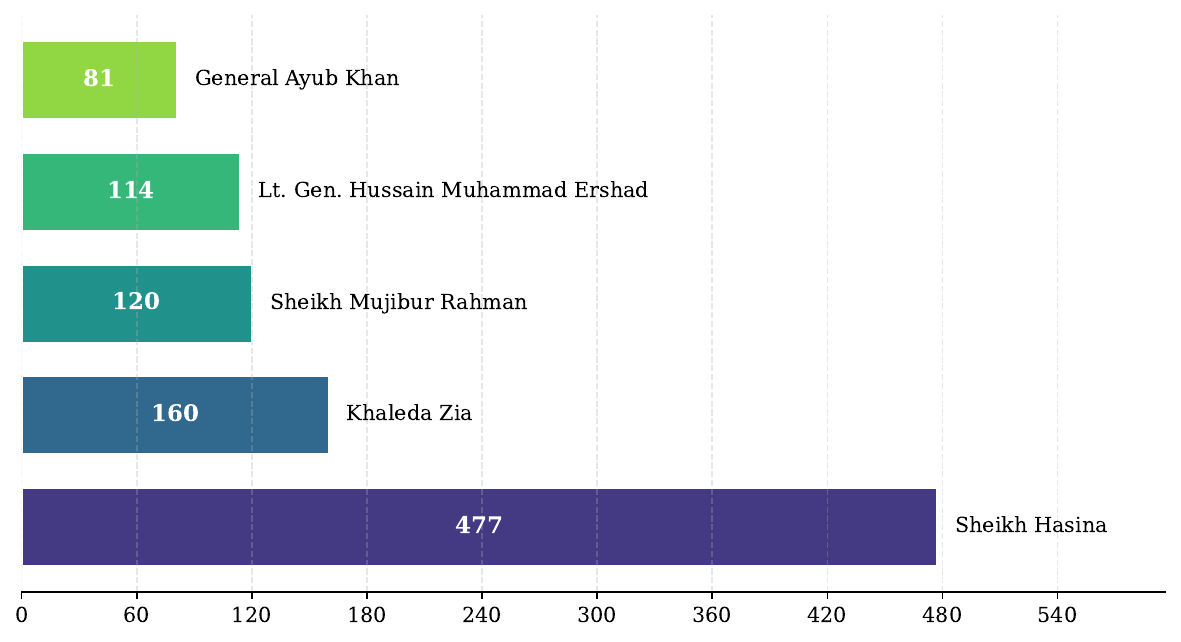}
    \caption{Top five heads of state by legislative activity.}
    \label{fig:top-heads}
\end{figure}

The corpus also permits analysis of output by individual leaders (Figure~\ref{fig:top-heads}).

\begin{itemize}
    \item \textbf{Sheikh Hasina:} the most prolific, with 477 acts (32.1\% of the total), reflecting extended tenure across multiple terms.
    \item \textbf{Khaleda Zia:} the second most prolific, with 160 acts.
    \item \textbf{Sheikh Mujibur Rahman:} 120 acts, with the highest annual rate at roughly 40 acts per year during a brief tenure (1972--1975).
    \item \textbf{Military leaders} (e.g., Lt.\ Gen.\ Hussain Muhammad Ershad, General Ayub Khan): substantial output concentrated on administrative and developmental policy.
\end{itemize}

These figures illustrate the influence of individual leadership on the composition of the statutory corpus.

\subsection{Legal Complexity and Scope}
Acts vary considerably in structural complexity, measured by section count:
\begin{itemize}
    \item \textbf{The Code of Criminal Procedure (1898):} the most extensive act, with 586 sections.
    \item \textbf{The Penal Code (1860):} 583 sections.
    \item \textbf{The Bangladesh Merchant Shipping Ordinance (1983):} 516 sections, indicating comparable complexity in modern legislation.
    \item \textbf{Recent acts} such as the Companies Act (1994) and the Bangladesh Labour Act (2006) retain substantial complexity, with 418 and 355 sections respectively.
\end{itemize}

Structural complexity is therefore not confined to any single period; substantial acts appear across the full temporal span.

\subsection{Active and Repealed Acts}
\begin{figure}[!htb]
    \centering
    \includegraphics[width=.7\linewidth]{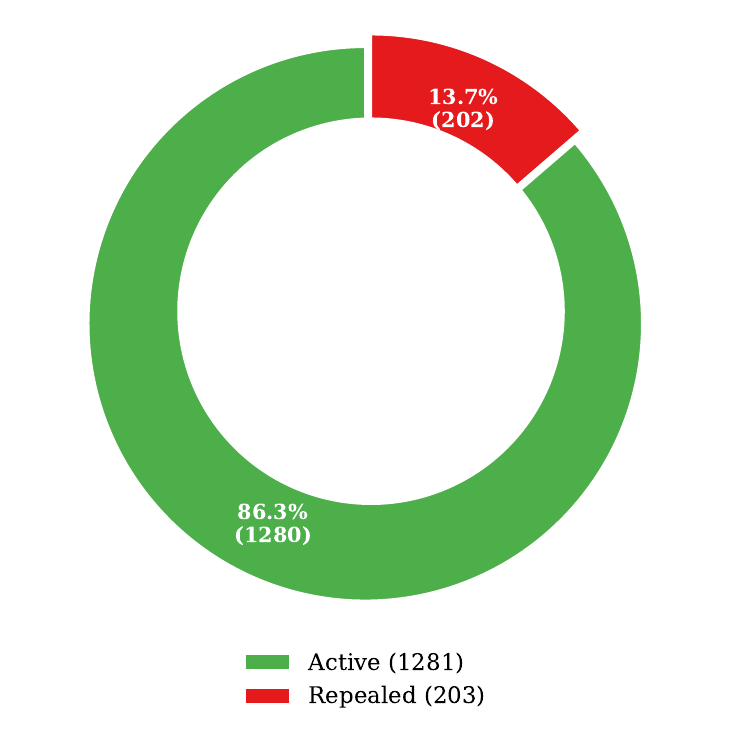}
    \caption{Active versus repealed acts.}
    \label{fig:active-repealed}
\end{figure}

Legal systems evolve in part by repealing acts that are no longer required, keeping the statutory framework current (Figure~\ref{fig:active-repealed}).

Of the acts in the corpus, 1{,}280 are currently active and 203 have been repealed.

\subsection{Repeals and Legal Evolution}
Repeal patterns offer a view of how legislation is revisited across administrations:
\begin{itemize}
    \item \textbf{Lt.\ Gen.\ Hussain Muhammad Ershad:} 114 acts enacted during his tenure, of which 62 were subsequently repealed.
    \item \textbf{Ziaur Rahman:} 70 acts enacted, of which 31 were subsequently repealed.
    \item \textbf{Other leaders:} additional acts were repealed after the enacting administration left office, consistent with ongoing legal revision.
\end{itemize}

These patterns are consistent with the routine revision of statutory law as political leadership changes, and provide a basis for further study of legal evolution.

\section{Use Cases}

BLAD can serve as a foundation for research in multilingual NLP, legal history, and governance. Rather than prescribing a fixed agenda, we outline several directions to indicate the corpus's potential and to invite further work. In each case we note the validation that responsible use would require.

\subsection{Legal Text Classification}
With 1{,}484 acts spanning multiple regimes, the corpus supports experiments in automatic categorisation by subject, legal system, or period. Careful validation remains necessary, but such models could assist document retrieval and comparative analysis.

\subsection{Temporal Analysis}
Because the corpus spans more than two centuries, it enables exploratory study of how legal content and structure change over time---for instance the emergence of new concepts or shifts in document length---subject to caveats about archival gaps and historical context.

\subsection{Retrieval-Augmented Generation (RAG)}
The corpus can support prototype question-answering systems that retrieve relevant passages before generating a response. Initial work could target fact-based queries (e.g., reforms enacted during the 1970s), with outputs evaluated against expert judgement.

\subsection{Historical Governance Studies}
Linking acts to political timelines allows analysis of how successive administrations structured the law. Any causal claim would require a robust methodology and additional socio-political data.

\subsection{Multilingual NLP}
Because statutes appear in Bengali, English, and mixed script, the corpus offers test material for tokenisation, named-entity recognition, and domain-specific machine translation, complementing larger multilingual resources.

\subsection{Fine-tuning Legal Language Models}
Fine-tuning existing language models on the corpus may improve their handling of Bangladeshi legal prose, with initial evaluation on tasks such as clause summarisation or terminology explanation.

\subsection{Policy Impact Exploration}
The corpus can support descriptive studies relating legislation to socio-economic indicators. Predictive work is feasible but would benefit from interdisciplinary collaboration and transparent error analysis.

\section*{Limitations}

While BLAD is a historically and structurally rich resource, several limitations affect its completeness and downstream applicability.

First, its legal scope is restricted to legislative acts and excludes case law, judicial interpretations, executive orders, and policy documents. This restricts tasks such as judgment prediction and case matching, which require multi-source legal corpora.

Second, the multilingual nature of Bangladeshi legislation means the corpus contains Bengali, English, and mixed-language documents. Although language identification has been applied, residual errors are likely, particularly for short or code-switched sections; transformer-based multilingual tokenisation (e.g., XLM-R) is a candidate for future improvement.

Third, the contextual metadata---political regime, head of state, and government structure---is inferred from secondary sources and historical mapping rather than encoded in the acts themselves. Such mappings may contain temporal ambiguities, especially for transitional regimes, and would benefit from expert-validated annotation in future releases.

Fourth, quality assurance combined rule-based validation with manual spot-checks, but comprehensive human verification remains pending. Statistical audits confirmed structural consistency, yet the correctness of named-entity mentions, token counts, and annotations would be strengthened by community review.

Finally, the corpus does not yet include downstream benchmarking or integration with established legal NLP pipelines (e.g., LexGLUE \cite{chalkidis_lexglue_2022}). The use cases outlined above---retrieval-augmented generation, temporal analysis, and bilingual search---are proposed directions; empirical demonstrations remain future work.

We regard these limitations as opportunities for community-driven improvement. The dataset is released under CC~BY-SA~4.0, and we welcome contributions that extend coverage, validate metadata, and improve interoperability with multilingual legal tools.

\bibliography{BlAD}
\bibliographystyle{acl_natbib}

\appendix

\section{Dataset Schema}
\label{sec:appendix-schema}

Each entry in BLAD is stored as a structured JSON object. The principal fields are summarised below:

\begin{itemize}
    \item \textbf{act\_title}: full title of the act.
    \item \textbf{act\_no}: act number (e.g., ``V'').
    \item \textbf{act\_year}: year of enactment.
    \item \textbf{sections}: list of sections with content text.
    \item \textbf{footnotes}: list of footnotes with amendment history.
    \item \textbf{language}: detected language of the act (Bengali, English, or mixed).
    \item \textbf{token\_count}: number of tokens after cleaning and standardisation.
    \item \textbf{repealed}: boolean indicating whether the act is repealed.
    \item \textbf{government\_context}: metadata on the governing regime at enactment.
    \item \textbf{legal\_system\_context}: legal framework, court structure, policing, and land relations.
    \item \textbf{processing\_info}: timestamps and enhancements applied during cleaning.
    \item \textbf{source\_url}: URL from which the act was originally retrieved.
    \item \textbf{copyright\_info}: attribution to the Ministry of Law.
\end{itemize}

\section{Sample JSON Entry}
\label{sec:appendix-sample}

A representative record for the act titled \textit{THE [***] WILLS AND INTESTACY REGULATION, 1799} is shown in Figure~\ref{fig:sample-json}; for space, only selected fields are included.

\begin{figure*}[!t]
\centering
\fbox{%
\begin{minipage}{0.95\textwidth}
\footnotesize
\ttfamily
\obeylines
\noindent
\{
  "act\_title": "THE [***] WILLS AND INTESTACY REGULATION, 1799",\\
  "act\_no": "V",\\
  "act\_year": 1799,\\
  "language": "english",\\
  "token\_count": 1277,\\
  "sections": [\\
  \hspace*{1em}\{"section\_content": "2. In all cases of Hindu, Mussalman ..."\},\\
  \hspace*{1em}\{"section\_content": "3. In case of a Hindu, Mussalman ..."\},\\
  \hspace*{1em}...\\
  ],\\
  "footnotes": [\\
  \hspace*{1em}\{"footnote\_text": "1The word 'Bengal' was omitted ..."\},\\
  \hspace*{1em}\{"footnote\_text": "2The word 'Bangladesh' was substituted ..."\}\\
  ],\\
  "government\_context": \{\\
  \hspace*{1em}"govt\_system": "Company Rule (Dual System until 1772)",\\
  \hspace*{1em}"head\_govt\_name": "Marquess of Wellesley",\\
  \hspace*{1em}"period\_years": "1799--1805"\\
  \},\\
  "legal\_system\_context": \{\\
  \hspace*{1em}"period\_info": \{"period\_name": "East India Company Rule"\},\\
  \hspace*{1em}"legal\_framework": \{\\
  \hspace*{2em}"primary\_laws": [\\
  \hspace*{3em}"Regulating Act 1773",\\
  \hspace*{3em}"Cornwallis Code 1793"\\
  \hspace*{2em}],\\
  \hspace*{2em}"legal\_basis":\\
  \hspace*{3em}"Hybrid of British common law and Islamic/Hindu law"\\
  \hspace*{1em}\}\\
  \},\\
  "source\_url":\\
  \hspace*{1em}"http://bdlaws.minlaw.gov.bd/act-print-1315.html"\\
\}
\end{minipage}%
}
\caption{A representative BLAD JSON record, showing selected fields for \textit{THE [***] WILLS AND INTESTACY REGULATION, 1799}.}
\label{fig:sample-json}
\end{figure*}

\section{Dataset License and Terms of Use}
\label{sec:appendix-license}

The dataset is released under the Creative Commons Attribution-ShareAlike 4.0 International license (CC~BY-SA~4.0). Users may share and adapt the material for any purpose, including commercial use, provided that appropriate attribution is given and that derivative works are distributed under the same license. The underlying legal texts originate from the official Bangladesh Laws Portal and remain subject to attribution to the Ministry of Law, Justice and Parliamentary Affairs. BLAD is an unofficial compilation intended for research; for authoritative legal reference, users should consult the official portal.

\section{Extended Charts and Tables}
\label{sec:appendix-extended}

This appendix collects the full-resolution charts and supplementary tables abbreviated in the main text, including per-decade counts, per-government-system breakdowns, and repeal statistics by head of state.

\end{document}